\documentclass{article}

\usepackage[preprint,nonatbib]{neurips_2020}
\usepackage[utf8]{inputenc} 
\usepackage[T1]{fontenc}    
\usepackage{hyperref}       
\usepackage{url}            
\usepackage{booktabs}       
\usepackage{amsfonts}       
\usepackage{nicefrac}       
\usepackage{microtype}      

\usepackage{graphicx}
\usepackage{subfigure}
\usepackage{wrapfig}
\usepackage{amsmath}
\usepackage{amsthm}
\usepackage[ruled,vlined]{algorithm2e}
\usepackage[skip=3pt]{caption}
\usepackage{todonotes}

\newtheorem{theorem}{Theorem}

\newcommand{\widebar}[1]{\mkern5mu\overline{\mkern-5mu#1\mkern-5mu}\mkern5mu}
\newcommand{\R}{\mathbb{R}}
\newcommand{\E}{\mathbb{E}}
\newcommand{\KL}{\text{KL}}
\DeclareMathOperator{\argmax}{argmax}
\DeclareMathOperator{\softmax}{softmax}

\title{Dynamic Feature Acquisition with Arbitrary Conditional Flows}

\author{%
  Yang Li\\
  Department of Computer Science\\
  University of North Carolina at Chapel Hill\\
  \texttt{yangli95@cs.unc.edu} \\
  \And
  Junier B. Oliva \\
  Department of Computer Science\\
  University of North Carolina at Chapel Hill\\
  \texttt{joliva@cs.unc.edu}
}

\begin{document}

\maketitle

\begin{abstract}
Many real-world situations allow for the acquisition of additional relevant information when making an assessment with limited or uncertain data. However, traditional ML approaches either require all features to be acquired beforehand or regard part of them as missing data that cannot be acquired. In this work, we propose models that dynamically acquire new features to further improve the prediction assessment. To trade off the improvement with the cost of acquisition, we leverage an information theoretic metric, conditional mutual information, to select the most informative feature to acquire. We leverage a generative model, arbitrary conditional flow (ACFlow), to learn the arbitrary conditional distributions required for estimating the information metric. We also learn a Bayesian network to accelerate the acquisition process. Our model demonstrates superior performance over baselines evaluated in multiple settings\footnote{please refer to \url{https://arxiv.org/abs/2010.02433} for the latest version.}.
\end{abstract}

\section{Introduction}

A typical machine learning paradigm for discriminative tasks is to learn the distribution of an output, $y$ given a complete set of features, $x \in \R^d$: $p(y \mid x)$. Although this paradigm is successful in a multitude of domains, it is incongruous with the expectations of many real-world intelligent systems in two key ways: first, it assumes that a complete set of features has been observed; second, as a consequence, it also assumes that no additional information (features) of an instance may be obtained at evaluation time. These assumptions often do not hold; human agents routinely reason over instances with incomplete data and decide when and what additional information to obtain. Take for instance a doctor making a diagnosis on a patient. The doctor usually has not observed all the possible measurements (such as blood samples, x-rays, etc.) for the patient. He/she is also not forced to make a diagnosis based on the observed measurements; instead, he/she may dynamically decide to take more measurements to help determine the diagnosis. Of course, the next measurement to make (feature to observe), if any, will depend on the values of the already observed features; thus, the doctor may determine a different set of features to observe from patient to patient (instance to instance) depending on the values of the features that were observed. For example, a low value from a blood test may lead a doctor to ask for a biopsy, whereas a high value may lead to an MRI. Hence, not each patient will have the same subset of features selected (as would be the case with typical feature selection). 
In order to more closely match the needs of many real-world applications, we propose a dynamic feature acquisition model that not only makes predictions with incomplete/missing features, but also determines what next feature would be the most valuable to obtain for a particular instance.

Our proposed model is deployed in an $\R^d$ feature space, where we have a subset (perhaps empty) of observed features $o \subseteq \{1,\ldots,d\}$ with values $x_o \in \R^{|o|}$ and we can acquire new features from the unobserved subset $u = \{1,\ldots,d\} \setminus o$. To simplify the problem, we utilize a sequential acquisition strategy, where only one feature is acquired at each step. There are two desired properties for acquiring a new feature: the feature should be informative for the target variable $y$; the feature contains non-redundant information which cannot be inferred from $x_o$. We leverage the conditional mutual information, $I(x_i;y \mid x_o)$, to quantify the dependency between the target variable $y$ and each candidate feature $i \in u$. The feature with maximum mutual information will be the next feature to acquire. Before we actually perform the acquisition, we do not know the exact value of $x_i$. After acquiring its value, we add the newly acquired feature $i$ to the observed subset and proceed to the next acquisition step if necessary. 

Estimating mutual information in general is non-trivial \cite{kraskov2004estimating}. We need to capture the joint distribution, $p(x_i, y \mid x_o)$, and the marginal distributions $p(x_i \mid x_o)$ and $p(y \mid x_o)$. Our paradigm introduces another level of complexity. Since we require our model to be able to reason regardless of the set of previously observed features, both $x_o$ and $x_u$ could be arbitrary subsets. Not only does this mean the dimensionality of $x_o$ and $x_u$ can be arbitrary, but the type of dependencies we need to capture also scales exponentially. A naive method where different models $p_o(\cdot \mid x_o)$ are built for each different conditioning $x_o$ will quickly fail since it requires an exponential number of models with respect to the dimensionality $d$. In this work, we leverage a recently proposed generative model, ACFlow \cite{li2019flow}, to learn the arbitrary conditional distributions with one unified model.

In summary, our model sequentially selects features to acquire their actual values and extend the observed feature subset accordingly until it reaches the stopping criterion (e.g., acquisition budget). Then, the model will make a final prediction using the current acquired feature subset.
Our model is different from conventional feature selection, where a \emph{fixed} subset of features is selected for all instances; in contrast,  our model selects a specialized subset for each instance \emph{dynamically}. 

Since the observed subset is updated as the acquisition process runs, we need to recalculate $I(x_i ; y \mid x_o)$ at each acquisition step. For an $\R^d$ feature space, each acquisition step has $O(d)$ time complexity. We further propose to leverage the Bayesian network structure over $x$ and $y$ to reduce the search space. We also propose a modified Grow-Shrink \cite{margaritis2000bayesian} algorithm to learn the Bayesian network structure for non-Gaussian data.

In addition to general real-valued data, we also apply our proposed method on time series, where the acquired features must follow the chronological order. Inspired by Thompson sampling \cite{thompson1933likelihood}, we integrate our knowledge about the order with a prior distribution and perform the acquisition based on the posterior. We also consider time-critical applications, where one may want to make a prediction as early as possible and acquire features consecutively but stop acquiring when prediction reaches a specified confidence threshold.

Our contributions are as follows: 1) We extend flow models, specifically ACFlow, to perform the dynamic feature acquisition task, which has not been explored previously. 2) We demonstrate the advantage of leveraging Bayesian network to reduce the searching space for feature acquisition. 3) We propose a modified Grow-Shrink algorithm to learn the Bayesian network structure from data, which leverages a deep generative model for conditional independence tests. 4) We propose a time series feature acquisition strategy to integrate our chronological prior knowledge. 5) We achieve state-of-the-art performance among information based dynamic acquisition methods on both synthetic and real-world datasets.

\section{Method}
In this section, we first formally describe the dynamic feature acquisition problem. Then, we introduce our method for acquiring new features. We describe our methods in detail for various settings. We also briefly review the ACFlow model \cite{li2019flow}, which we utilize to learn arbitrary conditional distributions. 

\subsection{Problem Formulation}
Consider a real-valued feature vector, $x \in \R^d$, and a target variable, $y$. We do not have access to all the entries of $x$. Instead, we only have an observed feature subset $o$ with values $x_o \in \R^{|o|}$, where $o \subseteq \{1,\ldots,d\}$ (may be empty), and we can acquire more feature values $x_u$, where $u = \{1,\ldots,d\} \setminus o$ to further improve our prediction. To simplify the problem, we acquire one feature $i$ at each acquisition step, where $i \in u$. After, the acquired feature $i$ is observed, added to the observed set, and one proceeds to the next acquisition step if necessary. The goal of dynamic feature acquisition is to acquire as few features as possible while predicting $y$ as accurately as possible. In next section, we describe our method to determine which feature, $i$, to acquire at each acquisition step. 

\subsection{Dynamic Feature Acquisition (DFA)}
In this section, we assume access to the arbitrary conditional distributions and leverage them for dynamic feature acquisition. Those conditional distributions can be estimated from a pretrained ACFlow model. We defer the training process to the next section.

The goal of feature acquisition is to increase the accuracy of our prediction, therefore, we would like the acquired feature to contain as much information about target variable as possible. We use the conditional mutual information to measure the amount of information. Specifically, we estimate the mutual information between each $x_i$ and $y$ conditioned on current observed set $x_o$, i.e., $I(x_i ; y \mid x_o)$. At each acquisition step, the feature with the highest mutual information will be our choice. Note that before we actually acquire the feature, we do not know the exact value of $x_i$, and we never observe the target variable $y$. Therefore, we need the arbitrary conditional distributions to infer these quantities. After the acquisition, the newly acquired feature, as well as its value, are added to the observed subset and the model proceeds to acquire the next feature. See Algorithm \ref{alg:dfa} for pseudo code of our acquisition approach.

\begin{algorithm}[t]
\SetAlgoLined
\KwIn{Training dataset $(\mathbf{X_t},Y_t)$ with all features and labels available; Test dataset $\mathbf{X_e}$ with no features available at the beginning; Total feature dimension $d$;}
1. Train ACFlow on training dataset by optimizing \eqref{eq:obj} \\
2. Dynamic feature acquisition and prediction: \\
\ForEach{$x \in \mathbf{X_e}$}{
    $o \leftarrow \emptyset; u \leftarrow \{1,\ldots,d\}; x_o \leftarrow \emptyset$ \tcp*[f]{initialize the observed set as empty}\\
    \Repeat{Stopping criterion reached}{
    estimate $I(x_i ; y \mid x_o)$ for $\forall i \in u$ \tcp*[f]{compute reward for each candidate}\\
    $i^* \leftarrow \argmax_i I(x_i ; y \mid x_o)$ \tcp*[f]{select the one with the highest reward}\\
    acquire the feature value $x_{i^*}$ \tcp*[f]{acquire the value of selected feature}\\
    $o \leftarrow o \cup i^*; u \leftarrow u \setminus i^*; x_o \leftarrow x_o \cup x_{i^*}$ \tcp*[f]{update observed and unobserved set}\\
    predict $y = \argmax p(y \mid x_o)$ \tcp*[f]{predict using current acquired subset}\\
    \tcp*[f]{stop if prediction is good enough or out of acquisition budget}
    }
}
\caption{Dynamic Feature Acquisition with ACFlow}
\label{alg:dfa}
\end{algorithm}

The conditional mutual information can be factorized as follows:
\begin{align}
    I(x_i ; y \mid x_o) &= I(x_i ; y, x_o) - I(x_i ; x_o) \label{cmi:0}\\
    &= H(x_i \mid x_o) - \E_{y \sim p(y \mid x_o)} H(x_i \mid y, x_o) \label{cmi:1}\\
    &= H(y \mid x_o) - \E_{x_i \sim p(x_i \mid x_o)} H(y \mid x_i, x_o) \label{cmi:2},
\end{align}
where $H(\cdot)$ represents the entropy. 
From \eqref{cmi:0}, we see our acquisition metric prefers features that contain more information about $y$ without redundant information already in $x_o$. 
According to \eqref{cmi:1}, when $x_o$ is observed, we prefer $x_i$ with higher entropy, that is, if we are already certain about the value of $x_i$, we do not need to acquire it anymore. Since $H(y \mid x_o)$ is fixed at each acquisition step, from \eqref{cmi:2}, we prefer feature that can reduce the most of our uncertainty about the target $y$.
In the following sections, we will describe how to estimate the conditional mutual information in different settings.

\subsubsection{Classification and Regression}
The real-valued target variable $y$ can be concatenated with $x$ as inputs to the ACFlow model so that $p(y \mid x_o)$ can be predicted by the same ACFlow as well.
The conditional mutual information, by definition, is
\begin{equation}\label{cmi:reg}
    I(x_i ; y \mid x_o) = \E_{p(x_i, y \mid x_o)} \log \frac{p(x_i, y \mid x_o)}{p(x_i \mid x_o)p(y \mid x_o)} = \E_{p(x_i, y \mid x_o)} \log \frac{p(y \mid x_i,x_o)}{p(y \mid x_o)}.
\end{equation}
We then perform a Monte Carlo estimation by sampling from $p(x_i, y \mid x_o)$. Note that $p(y \mid x_i, x_o)$ is evaluated on sampled $x_i$ rather than the exact value, since we have not acquired its value yet.

For a discrete target variable $y$, we employ Bayes's rule:
\begin{equation}\label{eq:bayes_rule}
    P(y \mid x_i,x_o) = \frac{p(x_i,x_o \mid y)P(y)}{\sum_{y'} p(x_i,x_o \mid y')P(y')} = \softmax_y(\log p(x_i,x_o \mid y') + \log P(y')).
\end{equation}
$P(y \mid x_o)$ can be similarly computed. The benefit of conditioning on $y$ is that we do not need an external classifier for prediction; we can use the same ACFlow model for both feature acquisition and prediction. To estimate conditional mutual information for classification problems, we can further simplify \eqref{cmi:reg} as
\begin{equation}\label{cmi:cls}
    I(x_i ; y \mid x_o) = \E_{p(x_i \mid x_o)P(y \mid x_i, x_o)} \log \frac{P(y \mid x_i,x_o)}{P(y \mid x_o)}
    = \E_{p(x_i \mid x_o)} \KL[P(y \mid x_i, x_o) \| P(y \mid x_o)],
\end{equation}
where the KL divergence between two discrete distributions can be analytically computed. Note $x_i$ is sampled from $p(x_i \mid x_o)$ as before. We again use Monte Carlo estimation for the expectation.

\subsubsection{Pruning the Searching Space}
\begin{wrapfigure}{r}{0.22\textwidth}
\vspace{-15pt}
\begin{center}
    \includegraphics[width=0.21\textwidth]{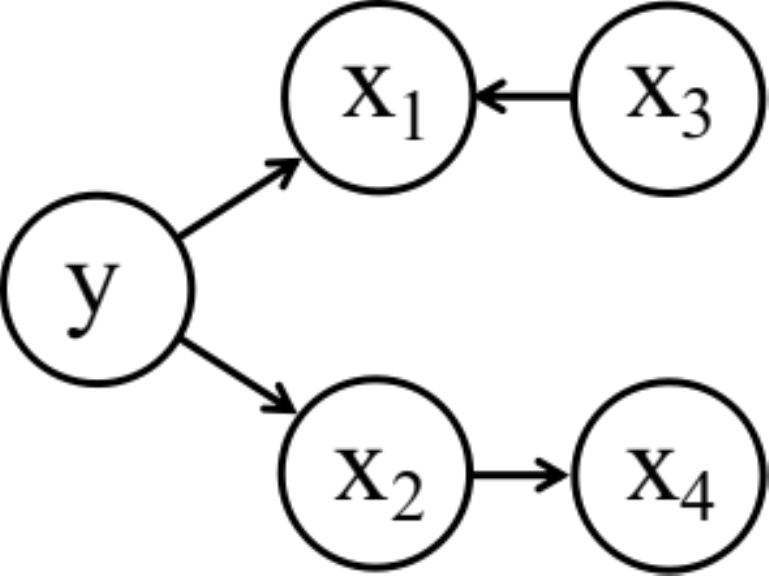}
\end{center}
\caption{An example of Bayesian Network.}
\label{fig:bn}
\end{wrapfigure}

As shown in Algorithm \ref{alg:dfa}, we need to estimate the conditional mutual information, $I(x_i ; y \mid x_o)$, for all possible candidates at each acquisition step. This induces a $O(d)$ time complexity at each acquisition step in $\R^d$ feature space. We propose to leverage the Bayesian Network to prune the searching space. Since our ultimate goal is to predict $y$, if observing $x_o$ makes part of $x_u$ independent of $y$, then we do not need to acquire those features. Note that acquiring new features might also introduce dependencies. For instance, in Fig.~\ref{fig:bn}, if we observe $x_2$, then $x_4$ is independent of $y$, therefore we do not need to acquire $x_4$ anymore; if we observe $x_1$, then $x_3$ becomes dependent, so we will need to reconsider $x_3$ as a candidate. 

This is different from using the Markov Blanket of $y$ for feature selection. For instance, in Fig.~\ref{fig:bn}, the Markov Blanket will constrain the searching space to $\{x_1, x_2, x_3\}$, and at each step a subset of the Markov Blanket is considered as candidates. However, in our method, the candidates are dynamically chosen; it can add or delete some features based on the current acquired features, which means our method potentially selects over fewer candidates.

\subsubsection{Bayesian Network Structure Learning}
As shown in the last section, the Bayesian network can be leveraged to reduce the searching space. However, the groundtruth BN is generally not known with real-world data. Therefore, we would like to learn the Bayesian network structure from data. Conventional methods typically consider either discrete or Gaussian distributed data. Here, we propose to utilize the dependencies captured by an ACFlow model to infer the Bayesian network in order to deal with non-Gaussian data.

\begin{wrapfigure}{r}{0.55\textwidth}
    \centering
    \vspace{-15pt}
    \subfigure[]{\includegraphics[width=0.11\textwidth]{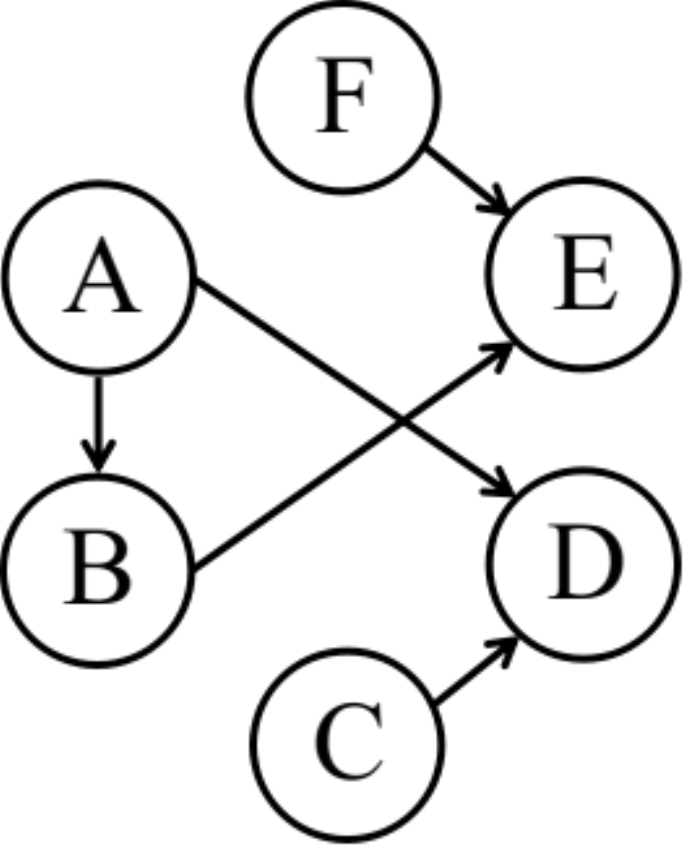}\label{fig:bnlearn:a}}
    \hspace{3mm}%
    \subfigure[]{\includegraphics[width=0.11\textwidth]{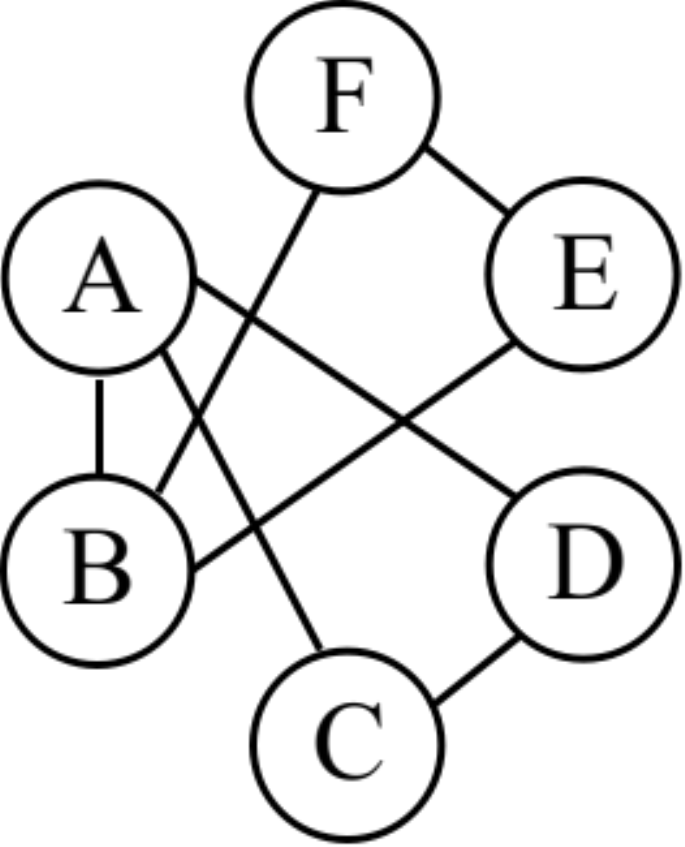}\label{fig:bnlearn:b}}
    \hspace{3mm}%
    \subfigure[]{\includegraphics[width=0.11\textwidth]{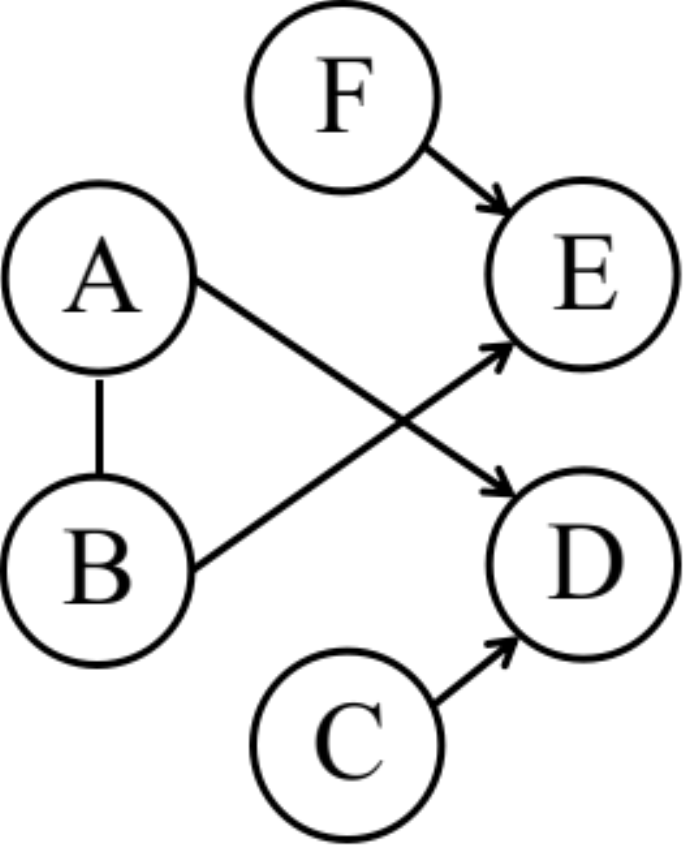}\label{fig:bnlearn:c}}
    \hspace{3mm}%
    \subfigure[]{\includegraphics[width=0.11\textwidth]{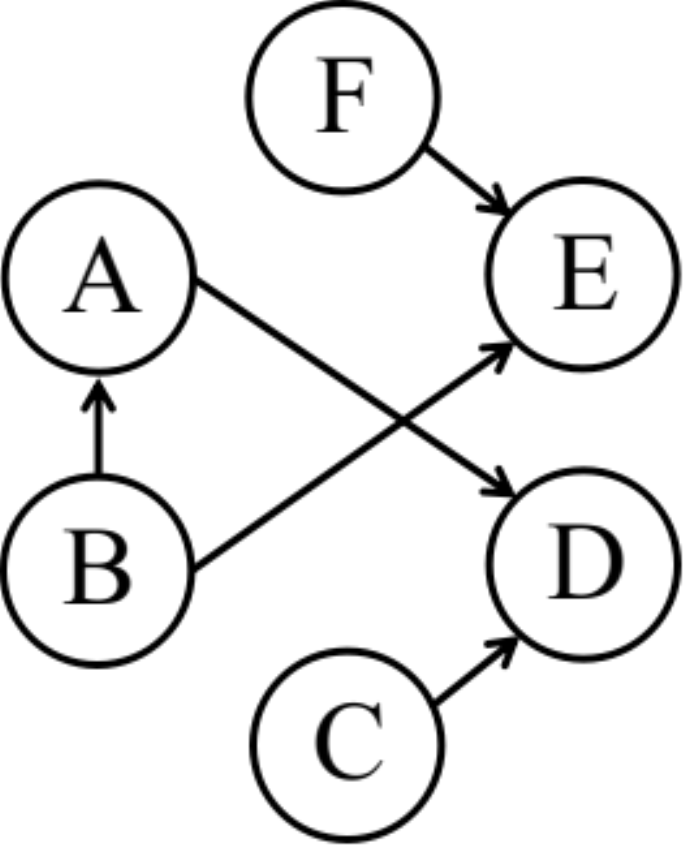}\label{fig:bnlearn:d}}
    \caption{
    Structure learning.
    (a) The groundtruth Bayesian network. (b) The moral graph where each node is linked to its Markov Blanket with undirected edges. (c) PDAG returned by GS algorithm. Note the edge between A and B is undirected. (d) Resolve the orientation for remaining undirected edges. Here, the orientation is randomly chosen, since both directions do not introduce v-structures.}
    \label{fig:bnlearn}
    \vspace{-7pt}
\end{wrapfigure}

Our method is based on the Grow-Shrink (GS) algorithm \cite{margaritis2000bayesian}. Grow-Shrink is a constraint based structure learning algorithm. It starts by identifying the Markov Blanket for each node (feature). After getting the Markov Blanket, we can build a moral graph \cite{lauritzen1988local}, where parents and children are linked with undirected edges, spouses are linked together as well (Fig.~\ref{fig:bnlearn:b}). Then conditional independence tests are performed to detect and delete spouse links. Edges involved in v-structures can further be oriented by performing conditional independence tests. The GS algorithm returns a partial directed acyclic graph (PDAG) where some edges cannot be oriented and remain undirected (Fig.~\ref{fig:bnlearn:c}).  We refer readers to \cite{margaritis2000bayesian,pellet2008using} for details and pseudo code of the GS algorithm. The final step is to orient the remaining undirected edges. Following \cite{chickering2002learning}, we orient those edges in a way that they will not introduce v-structures (Fig~\ref{fig:bnlearn:d}).

We first propose a way of identifying Markov Blanket for both features $x$ and target $y$ using ACFlow. For notation succinctness, we concatenate $y$ into $x$ and denote them as $\mathrm{x}$. Specifically, we demonstrate the following theorem (see Appendix~\ref{sec:proof} for proof):
\begin{theorem}\label{thm:bn}
Let $\mathrm{x} \in \mathcal{X}^d$ (either discrete or continuous or mixed), $MB(\cdot)$ represents the Markov Blanket of the input, $I(\mathrm{x}_i ; \mathrm{x}_j \mid \mathrm{x}_{\widebar{\{i,j\}}}) = 0 \iff \mathrm{x}_i \notin MB(\mathrm{x}_j) \land \mathrm{x}_j \notin MB(\mathrm{x}_i)$, where  $\mathrm{x}_{\widebar{\{i,j\}}}$ represents all other variables except $\mathrm{x}_i$ and $\mathrm{x}_j$.
\end{theorem}
The Markov Blanket for a specific feature $i$ can be identified by estimating the conditional mutual information $I(\mathrm{x}_i;\mathrm{x}_j \mid \mathrm{x}_{\widebar{\{i,j\}}})$ in a similar fashion as \eqref{cmi:reg} and \eqref{cmi:cls} for each $j \neq i$. If $I(\mathrm{x}_i;\mathrm{x}_j \mid \mathrm{x}_{\widebar{\{i,j\}}})$ is larger than a specified threshold $\epsilon$, we take $\mathrm{x}_i$ and $\mathrm{x}_j$ as conditionally dependent given $\mathrm{x}_{\widebar{\{i,j\}}}$. We hence add $\mathrm{x}_i$ and $\mathrm{x}_j$ to the Markov Blanket of $\mathrm{x}_j$ and $\mathrm{x}_i$ respectively.

Typical applications of GS algorithm assume that features are Gaussian distributed, therefore $\chi$-squared tests are used to test conditional independence. To alleviate the Gaussian assumption, we leverage the ACFlow model for these tests. We again threshold the conditional mutual information between corresponding variables to test the conditional independence. For instance, in order to test if $\mathrm{x}_i$ and $\mathrm{x}_j$ are independent conditioned on $\mathrm{x}_c$, we estimate $I(\mathrm{x}_i;\mathrm{x}_j \mid \mathrm{x}_c)$. If $I(\mathrm{x}_i;\mathrm{x}_j \mid \mathrm{x}_c) \leq \epsilon$, we regard them as conditionally independent and vice versa.

\vspace{-5pt}
\subsubsection{Time Series}\label{sec:ts}
\vspace{-5pt}
In this section, we apply our method on time series data. For example, consider a scenario where sensors are deployed in the field with very limited power. We would like the sensors to decide when to put themselves online to collect data. The goal is to make as few acquisitions as possible while still making an accurate prediction. In contrast to ordinary vector data, the acquired features must follow a chronological order, i.e., the newly acquired feature $i$ must occur after all elements of $o$ (since we may not go back in time to turn on sensors). In this case, it is detrimental to acquire a feature that occurs very late in an early acquisition step. For example, for a time series with total $T$ time steps, if we acquire feature $x_{T-1}$ as the first feature, then we will lose the opportunity to observe features ahead of it.

Inspired by Thompson sampling \cite{thompson1933likelihood,russo2017tutorial}, we employ a prior distribution to encode our chronological constraint. Specifically, we set the prior as a Dirichlet distribution that is biased towards the selection of earlier time steps:
$
    \pi(\rho) = \text{Dir}\left[\alpha(T- (\max(o)+1)),\ldots, \alpha(T-(T-1)) \right](\rho),
$
where $\alpha$ is a hyperparameter, $T$ is the total time steps, $\max(o)$ represents the latest time step already acquired, and $\rho$ is a distribution for acquisition over the remaining future time steps.
However, we still desire that the acquired features are informative for target $y$. Hence, we update the prior to a posterior using time steps $V$ that are drawn according to how informative they are: 
\begin{equation}
    p(V_n=t) \propto \exp(I(x_t ; y \mid x_o)),\ t \in \{\max(o)+1,\ldots, T-1 \},\ n \in \{1,\ldots,N\},
\end{equation}
where $N$ is the total number of samples.
Due to conjugacy, the posterior is also a Dirichlet distribution
\begin{equation}
    p(\rho \mid V) = \text{Dir}\left[\alpha(T- (\max(o)+1)) + \sum_{n=1}^{N} \mathbb{I}\{V_n=\max(o)+1\}, \ldots  \right](\rho).
\end{equation}
Samples from posterior represent the probabilities of choosing each candidate, which now prefer both earlier time steps and informative features. We draw a sample from posterior and select the most likely time step at each acquisition step.

In some cases, the number of sensor readings is not as important as making a prediction as early as possible. That is, sensors may be collecting data at each time step towards a prediction, but one would like to make a final prediction as soon as a confidence threshold is hit. This setting is especially applicable in time-critical applications such as an autonomous vehicle predicting if a nearby pedestrian will cross the street. In this setting, we propose to use the prediction probability, $\max_y p(y \mid x_o)$, as the stopping criterion. When the probability reaches the specified threshold, we stop acquiring more features (time steps) and predict the target variables. Note that deep models usually underestimate uncertainty \cite{gal2016uncertainty}. Therefore, the actual accuracy may be lower than the specified threshold. We perform uncertainty calibration following \cite{kumar2019verified}. Note we calibrate uncertainty for each time step separately by utilizing a held-out validation set.

\vspace{-8pt}
\subsection{Arbitrary Conditional Flow}
\vspace{-5pt}
Above we assumed access to the arbitrary conditional distributions via a pretrained ACFlow model \cite{li2019flow}. Here, we give a brief introduction to the ACFlow model and our modified training procedure in different settings.
ACFlow extends flow models to learn the arbitrary conditional distributions $p(x_u \mid x_o)$.
Since both $x_u$ and $x_o$ could be an arbitrary subset of $x$, one needs to model an exponential number of conditionals with respect to the dimensionality of $x$. \cite{li2019flow} utilizes a masking mechanism to train ACFlow in a multi-task fashion; i.e., \emph{all} conditionals are captured by a single model. The multi-task training mechanism allows the model to generalize to unseen combinations of $x_u$ and $x_o$. 

The original training process of ACFlow selects two non-overlapping subsets of features at random as $x_u$ and $x_o$ for each training example. Then, the training objective is to maximize the arbitrary conditional log likelihoods $\log p(x_u \mid x_o)$. In our case, we train ACFlow to optimize
\begin{equation}\label{eq:obj}
    \mathcal{L} = \log p(x_i, y \mid x_o) = \log p(x_i \mid x_o) + \log p(y \mid x_i, x_o),
\end{equation}
where $x_o$ is an arbitrary subset, while $x_i, i \in u$ is an arbitrary element in unobserved subset. Note that the actual value of $x_i$ is used during training.

For real-valued target variable, we concatenate $x$ and $y$ so that $p(x_i \mid x_o)$ and $p(y \mid x_i, x_o)$ can both be evaluated by ACFlow. 
For discrete target variable, we train a conditional version of ACFlow conditioned on $y$. Then $P(y \mid x_i, x_o)$ can be evaluated by Bayes rule as in \eqref{eq:bayes_rule}.  $p(x_i \mid x_o)$ can be computed by marginalizing out $y$:
$
    p(x_i \mid x_o) = \frac{\sum_{y'} p(x_i,x_o \mid y') P(y')}{\sum_{y'} p(x_o \mid y')P(y')},
$
where $p(x_i, x_o \mid y')$ and $p(x_o \mid y')$ are conditional likelihoods from the conditional ACFlow.

\vspace{-5pt}
\section{Related Works}
\vspace{-5pt}
\paragraph{Active Learning}
Traditional active learning \cite{settles2009active} performs instance level acquisition, where new instances are selected to acquire their labels to train a better model. Information based acquisition functions play an important role in this area \cite{mackay1992information,houlsby2011bayesian}. In contrast, DFA does not consider a labeling oracle at training time. Instead, we perform a feature level acquisition, where new features are acquired for the same instance at evaluation time. In addition, we require the model to make prediction based on partial observations, while active learning typically assumes all features are observed for selected instances.

\vspace{-7pt}
\paragraph{Feature Selection} Conventional methods for feature selection eliminate redundant features, which can help reduce computation complexity and improve generalization \cite{chandrashekar2014survey}. The same selected subset is then applied to each instance. In contrast, in DFA 
a personalized subset of features is selected for each instance depending on the specific values that observed features held. Thus, DFA allows for an instance specific strategy to select informative features. It is worth noting that DFA may be applied after an initial feature selection preprocessing step to reduce the search space. 

\vspace{-7pt}
\paragraph{Dynamic Feature Acquisition}
Acquiring features dynamically is of great use in many real-world decision-making scenarios. Several methods have been proposed to actively acquire new features in cost-sensitive setting \cite{chai2004test,ling2004decision,sheng2006feature,nan2014fast}. In \cite{zubek2002pruning,ruckstiess2011sequential}, the dynamic feature acquisition problem is cast as a Markov Decision Process. \cite{shim2018joint} further propose a deep reinforcement learning based method. As in active learning, using an information-based acquisition metric is a promising direction, but it imposes a requirement to learn the arbitrary conditional distributions. A recent work EDDI \cite{ma2018eddi} utilizes a VAE based generative model to learn arbitrary marginal distributions $p(x_o)$. Since the VAE assumes the likelihood $p(x_i \mid x_o)$ for each feature $x_i$ is conditionally independent, they hence evaluate $p(x_i \mid x_o)$ by $p(x_i \mid z)$, where $z$ is the latent code. In our method, we directly model the arbitrary conditional distributions $p(x_i \mid x_o)$ with a flow based generative model.

\vspace{-7pt}
\paragraph{Bayesian Network Structure Learning} Learning the Bayesian network structure has been studied for decades. There are two primary classes of algorithms: constraint-based algorithms, where conditional independence tests are utilized to infer the dependency structure, and score-based algorithms, which maximize goodness-of-fit scores \cite{scutari2018learns,scanagatta2019survey}. Hybrid methods that combine both approaches have also been considered \cite{scutari2018learns,scanagatta2019survey}. In this work, we extend the constraint-based Grow-Shrink algorithm \cite{margaritis2000bayesian} with deep generative models to exploit the benefits of better distribution modeling.

\vspace{-5pt}
\section{Experiments}
\vspace{-5pt}
In this section, we evaluate our model in different settings by comparing to EDDI \cite{ma2018eddi}, a state-of-the-art information based feature acquisition method. We evaluate two acquisition strategies: the dynamic feature acquisition (DFA), where specialized acquisition orders are applied for different instances, and the static feature acquisition (SFA), where the same acquisition order is applied for each instance. The static order is similarly decided by the conditional mutual information metric, but averaged over all test instances. For estimating the conditional mutual information, we draw 10 samples from the corresponding distributions. We also compare to a decision tree based approach, where multiple trees with different depth are fitted. The depth controls the number of acquired features.

\vspace{-5pt}
\paragraph{Classification and Regression}
\begin{figure}
    \vspace{-0.1in}
    \centering
    \subfigure[example of acquired features]{\includegraphics[width=0.52\textwidth]{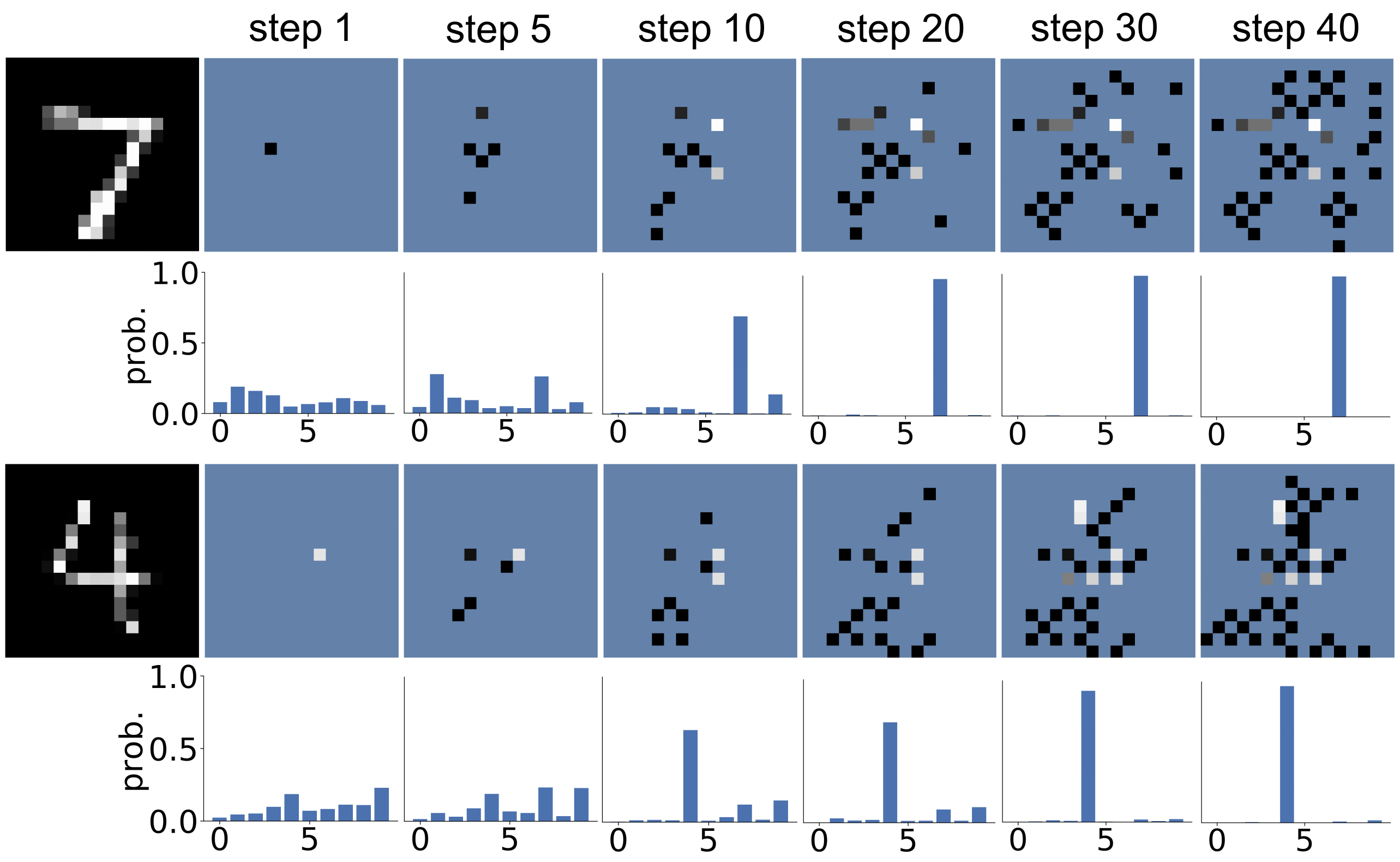}\label{fig:mnist:a}}
    \subfigure[test accuracy]{\includegraphics[width=0.4\textwidth]{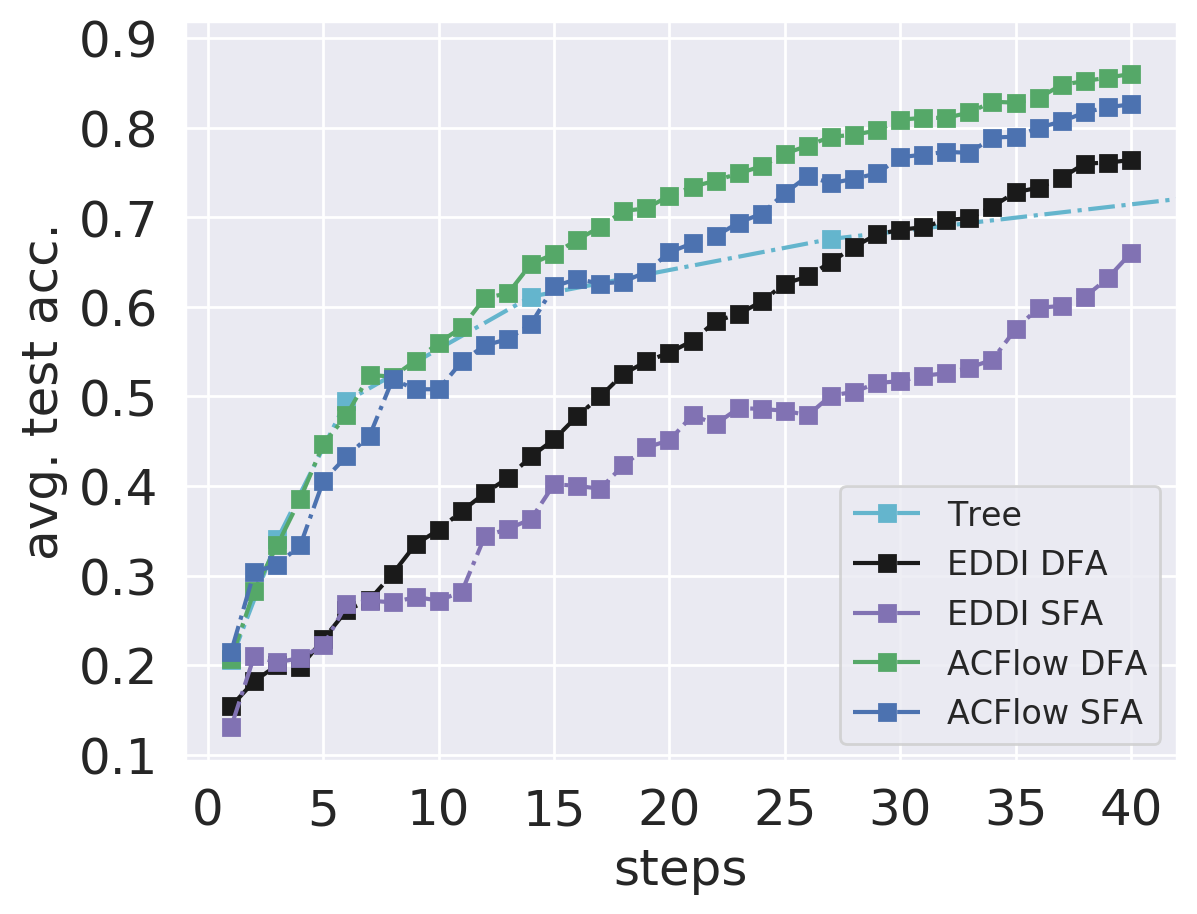}\label{fig:mnist:b}}
    \caption{Classification on MNIST. (a) two examples of the acquisition process. The blue mask indicates unobserved features. We also plot the prediction probability at each acquisition step. (b) prediction accuracy after each acquisition step averaged over the test set.}
    \label{fig:mnist}
    \vspace{-0.2in}
\end{figure}

\begin{figure}
    \begin{minipage}{0.48\textwidth}
    \centering
    \subfigure[synthetic]{\includegraphics[width=0.49\textwidth]{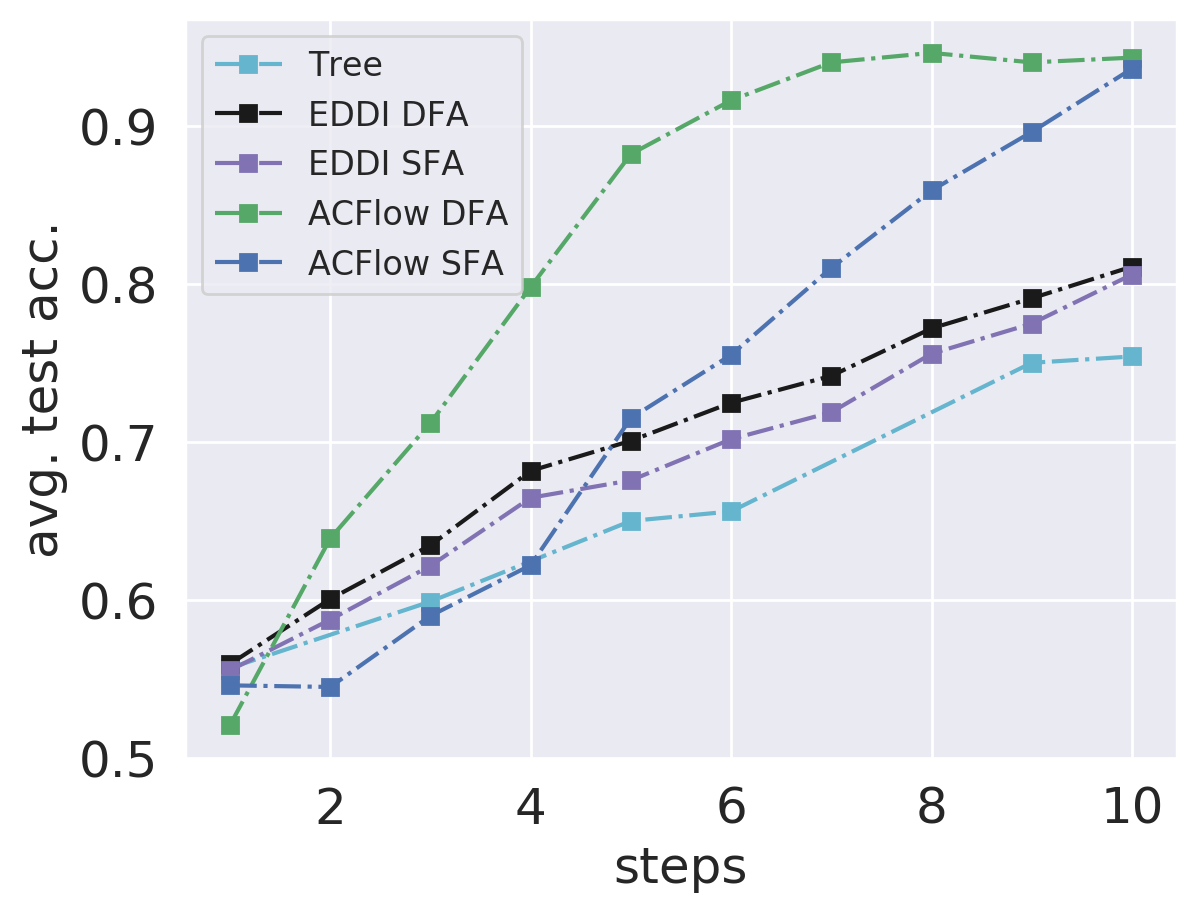}}
    \subfigure[gas]{\includegraphics[width=0.49\textwidth]{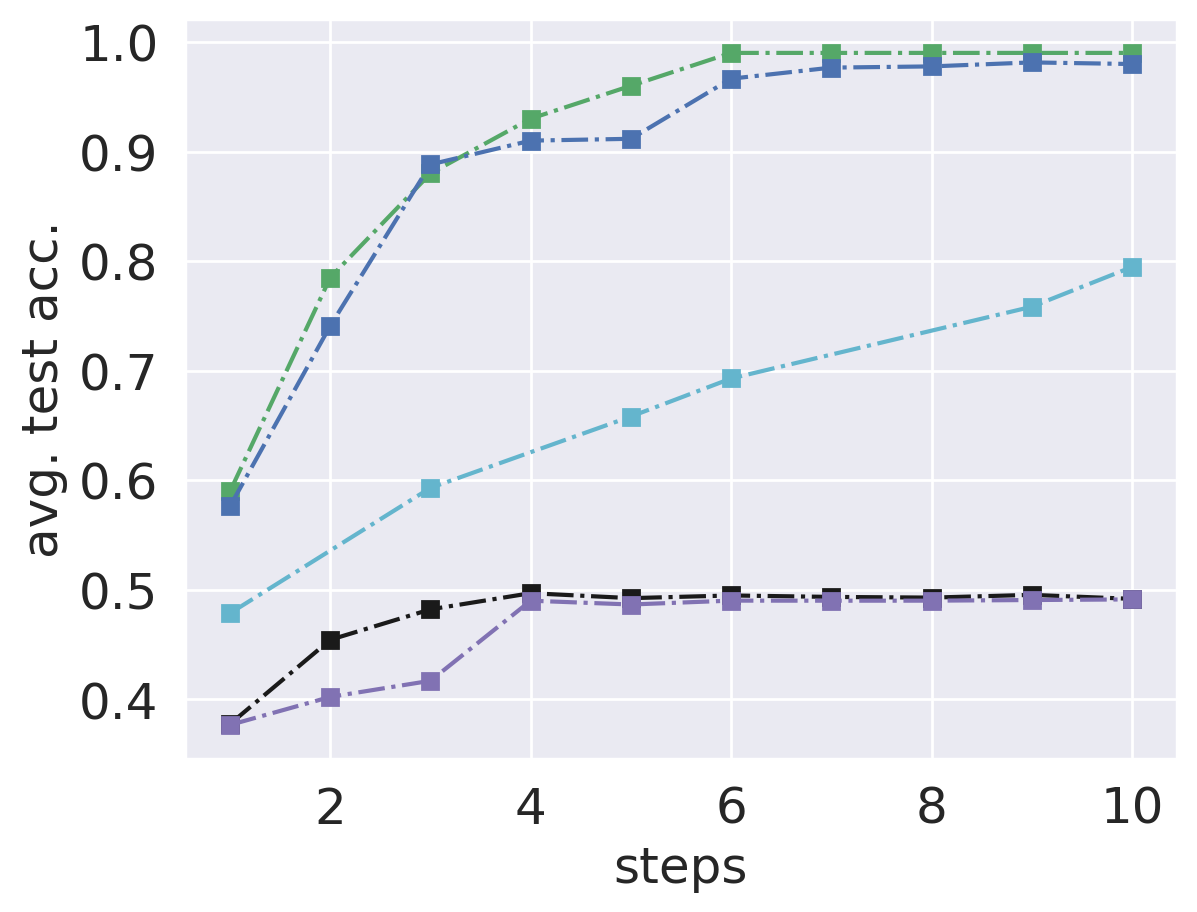}}
    \caption{Test Accuracy. Higher is better.}
    \label{fig:uci_cls}
    \end{minipage}
    \vspace{-0.1in}
    \hfill
    \begin{minipage}{0.5\textwidth}
    \centering
    \subfigure[housing]{\includegraphics[width=0.48\textwidth]{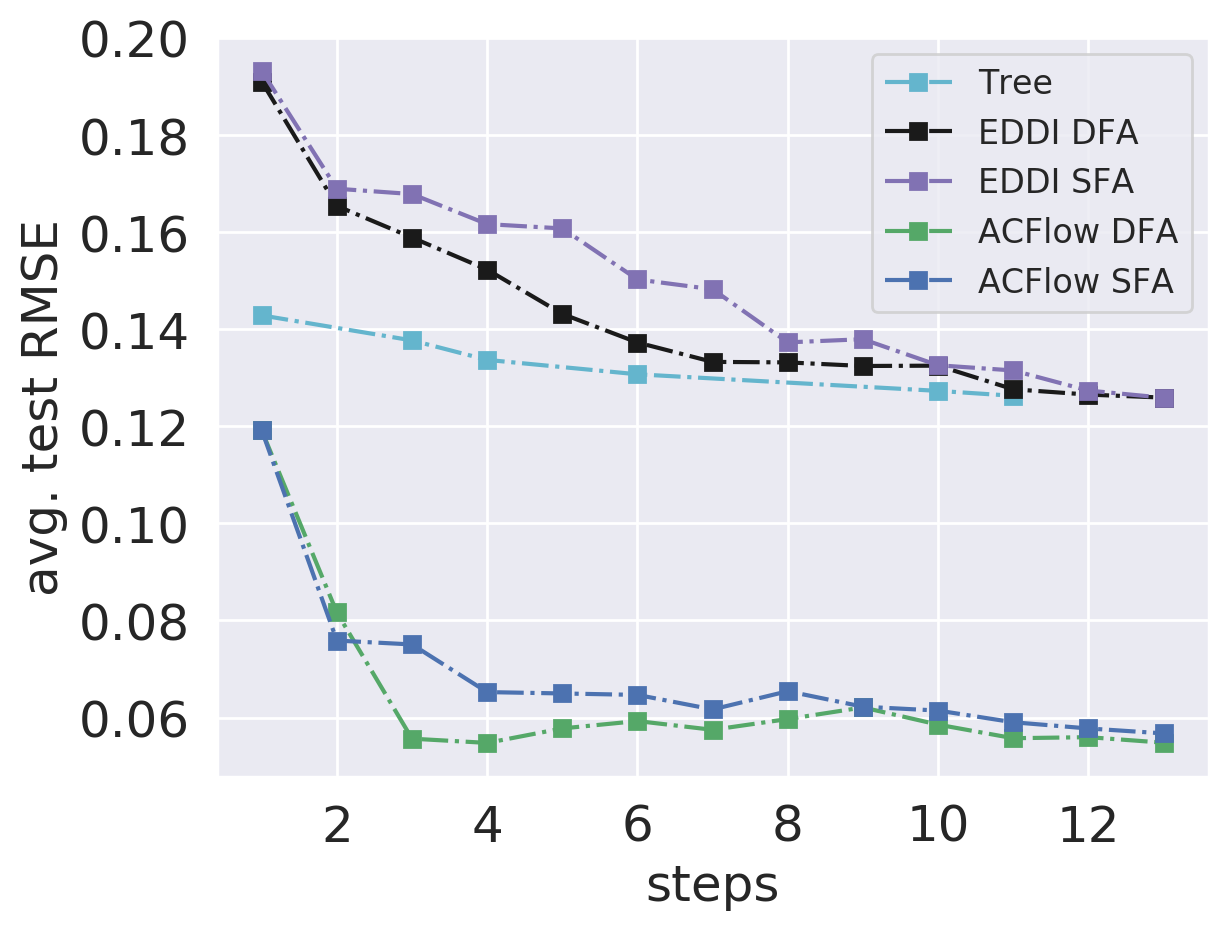}}
    \subfigure[whitewine]{\includegraphics[width=0.49\textwidth]{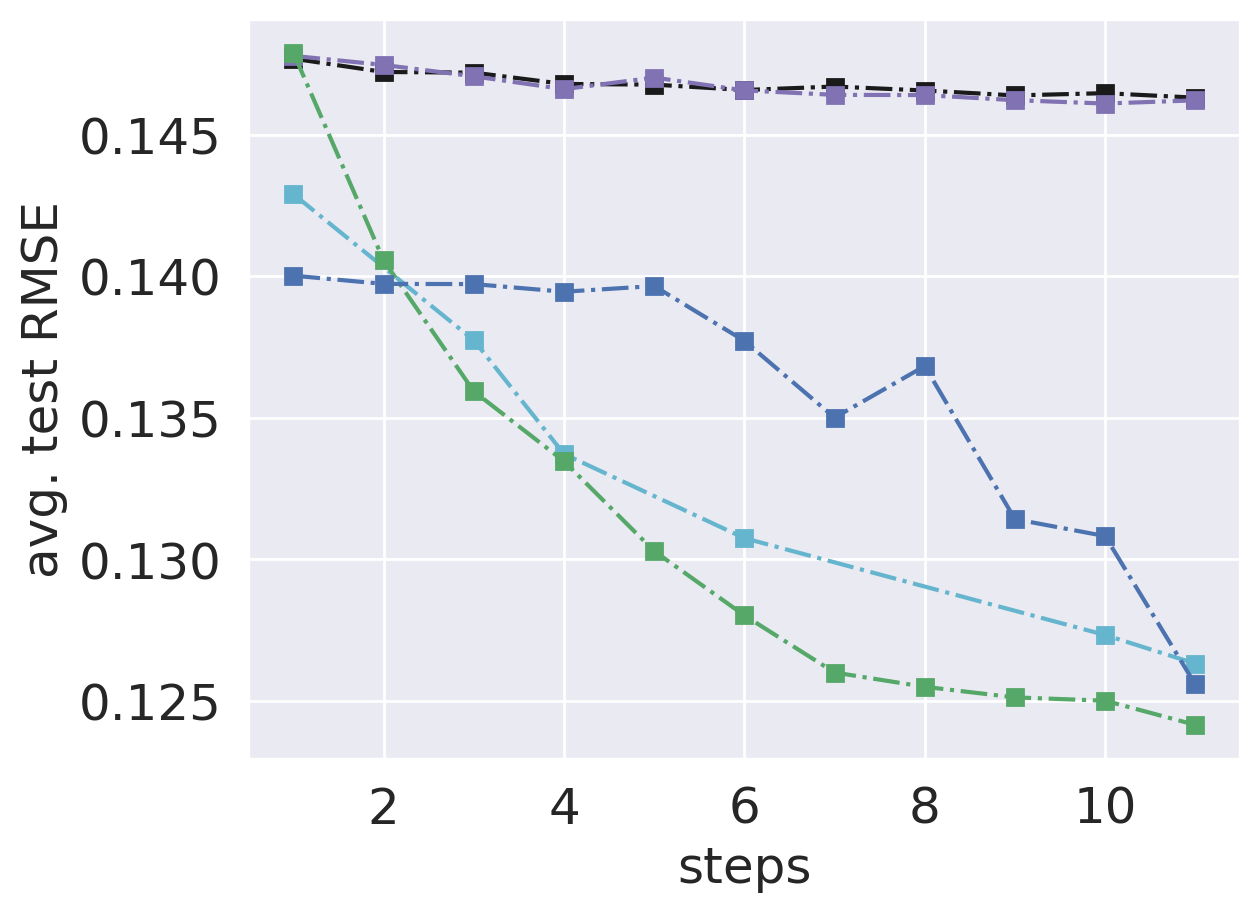}}
    \caption{Test RMSE. Lower is better.}
    \label{fig:uci_reg}
    \end{minipage}
    \vspace{-0.1in}
\end{figure}

In this section, we demonstrate that our model can achieve accurate classification and regression by acquiring only a small subset of features.
We first conduct experiments on MNIST dataset \cite{lecun1998gradient}. To reduce the number of total features, we downsample the images to $16 \times 16$. Data preprocessing and architecture details are listed in Appendix~\ref{sec:cls_reg}. The results as well as several examples are shown in Fig.~\ref{fig:mnist}. From Fig.~\ref{fig:mnist:a}, we can see our model acquires different features for different instances, which demonstrates that our model is capable of identifying specialized feature subsets for different instances. After acquiring only a small subset of features, the model is already certain about the target variable. It is also worth noting that our model is capable of exploring the spatial correlation between features. For instance, the acquired features present a checkerboard-like pattern, which is consistent with our impression that nearby pixels usually contain redundant information. Figure.~\ref{fig:mnist:b} compares the test accuracy after each acquisition step. Our model consistently outperforms EDDI at each acquisition step. We can see the DFA policy is always better than the SFA policy for both EDDI and ACFlow, which verifies the benefits of acquiring specialized subsets. We also notice that ACFlow with a static policy can already outperform EDDI with a dynamic policy, which we conjecture is due to the superior conditional distribution modeling ability of ACFlow. First, it improves the prediction performance since $p(y \mid x_o)$ is better captured by ACFlow. Second, it improves the conditional mutual information estimation, which gives better acquisition policy.

We then evaluate our model on real-valued vector data from both synthetic dataset and UCI datasets \cite{ucilichman}. Please refer to Appendix~\ref{sec:cls_reg} for experimental details. We employ a validation set to select the best architecture and hyperparameters for EDDI. Figures~\ref{fig:uci_cls} and~\ref{fig:uci_reg} present the results for classification and regression. We report the test accuracy and root mean squared error (RMSE) after each acquisition step for classification and regression tasks, respectively. We see our model achieves superior performance compared to EDDI. Additionally, the dynamic acquisition policy tends to give better results compared to a static alternative. Although the decision tree based method utilizes multiple models, our method exceeds it using only one unified model.

\begin{figure}
    \vspace{-0.1in}
    \begin{minipage}{0.48\textwidth}
    \centering
    \subfigure[asia]{\includegraphics[width=0.49\textwidth]{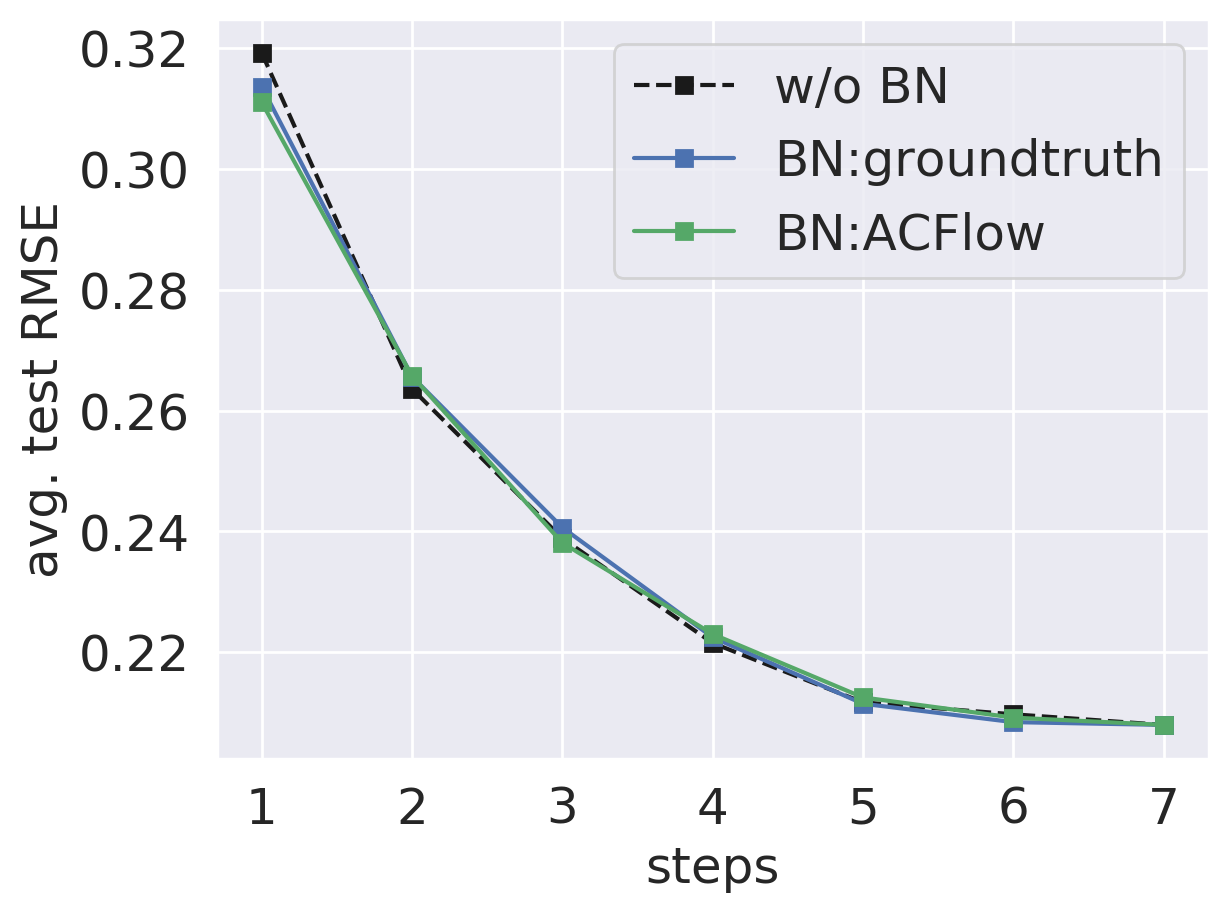}}
    \subfigure[sachs]{\includegraphics[width=0.49\textwidth]{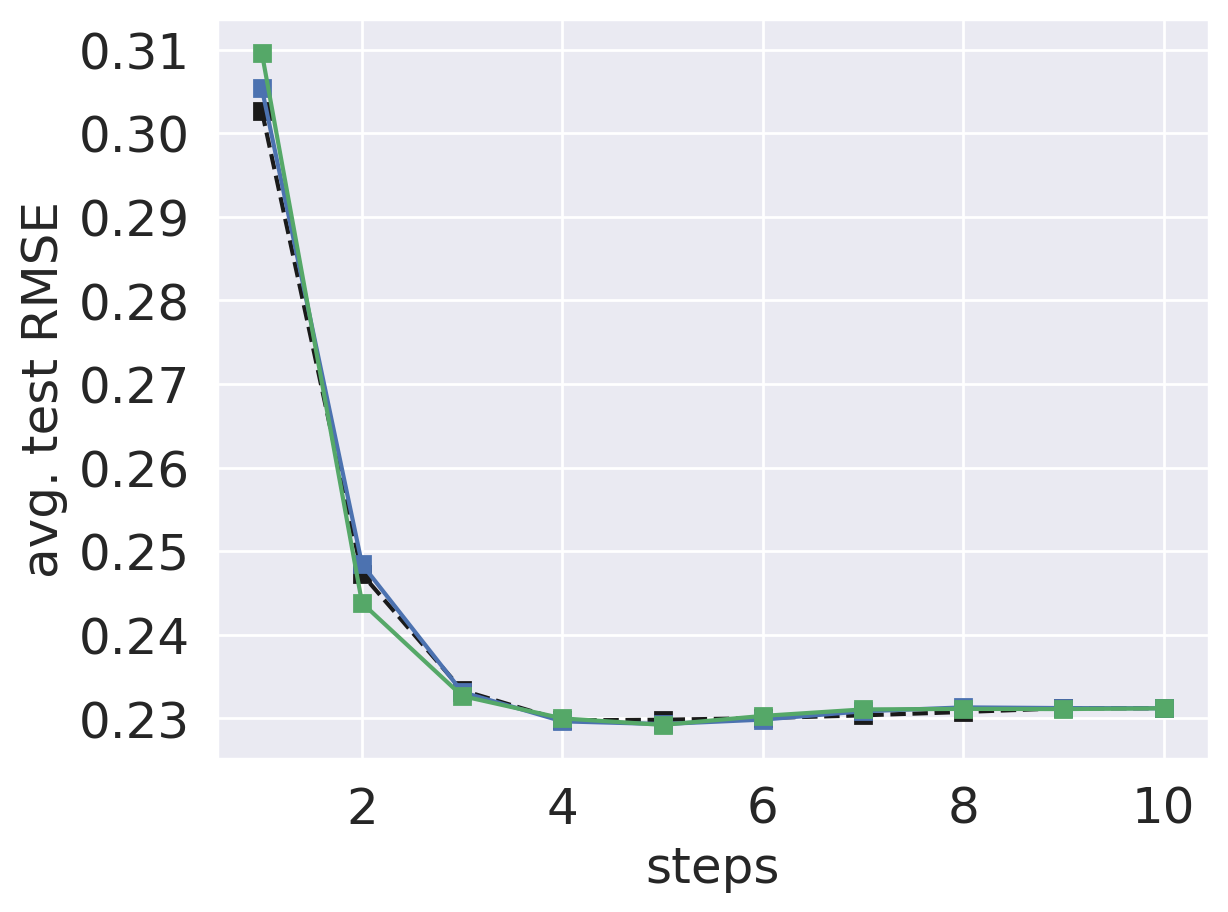}}
    \caption{On synthetic datasets, our learned BN performs equally well compared to groundtruth.}
    \label{fig:bn_syn}
    \end{minipage}
    \hfill
    \vspace{-0.1in}
    \begin{minipage}{0.48\textwidth}
    \centering
    \subfigure[housing]{\includegraphics[width=0.48\textwidth]{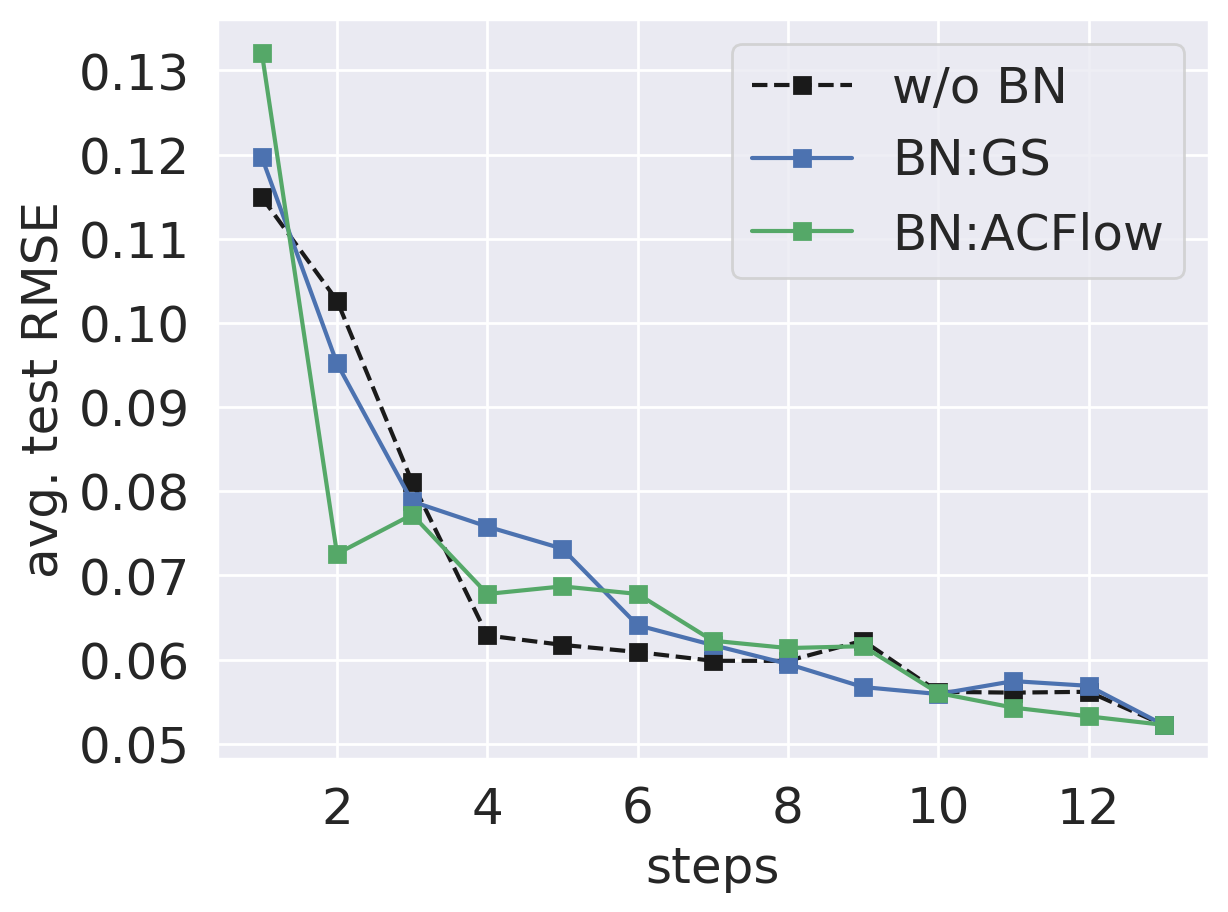}}
    \subfigure[whitewine]{\includegraphics[width=0.49\textwidth]{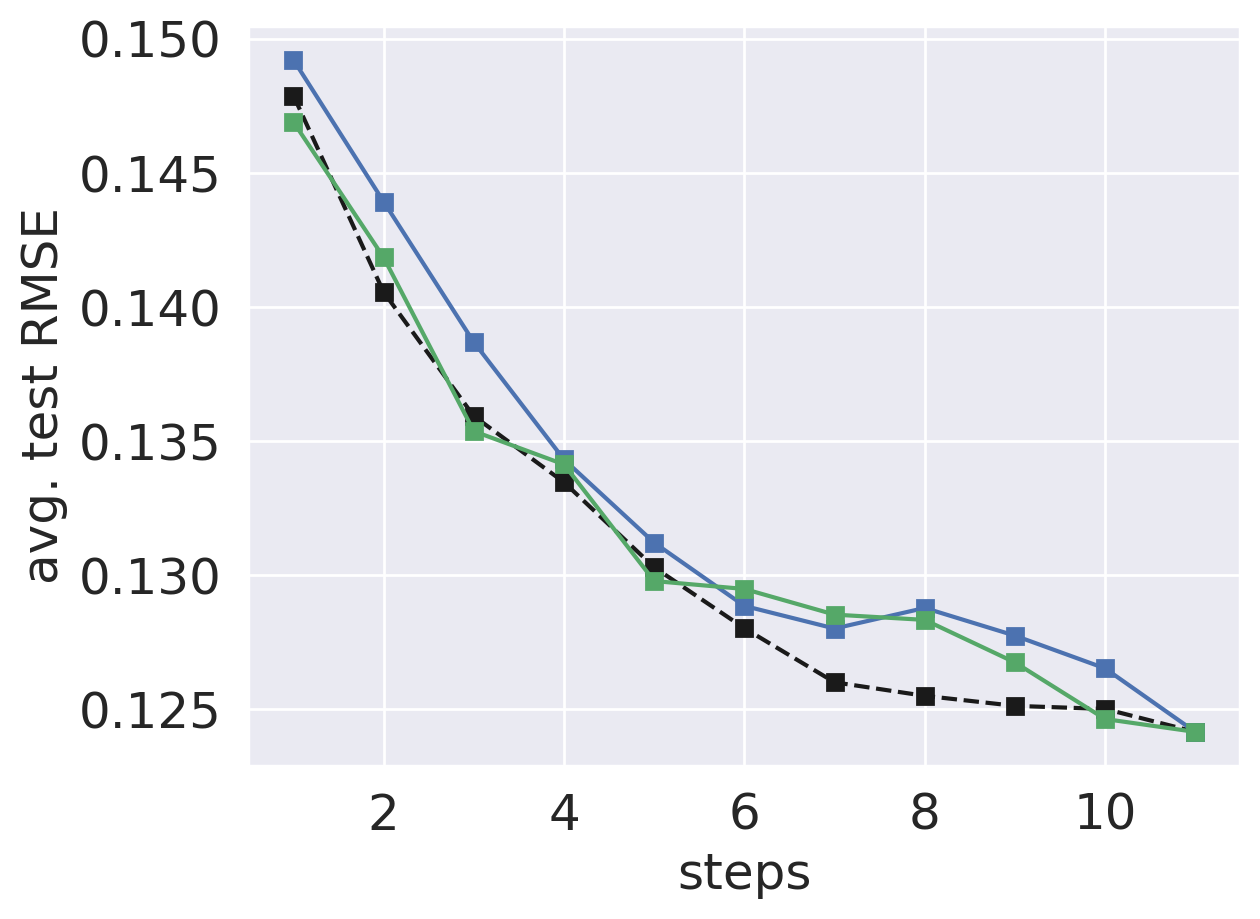}}
    \caption{On UCI datasets, our learned BN performs better than the one learned from GS.}
    \label{fig:bn_reg}
    \end{minipage}
\end{figure}

\begin{figure}
    \vspace{-0.1in}
    \begin{minipage}{0.48\textwidth}
    \centering
    \subfigure[digits]{\includegraphics[width=0.49\textwidth]{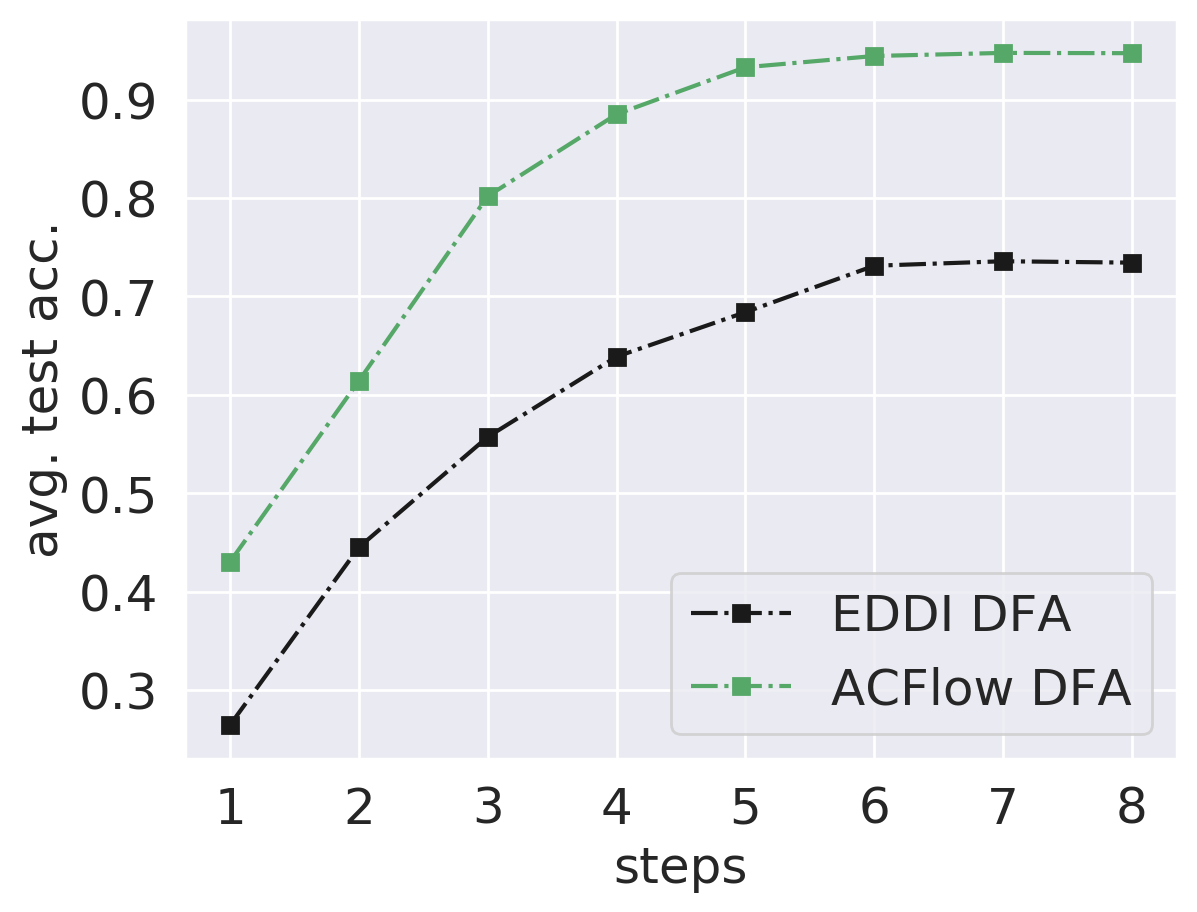}}
    \subfigure[pedestrian]{\includegraphics[width=0.49\textwidth]{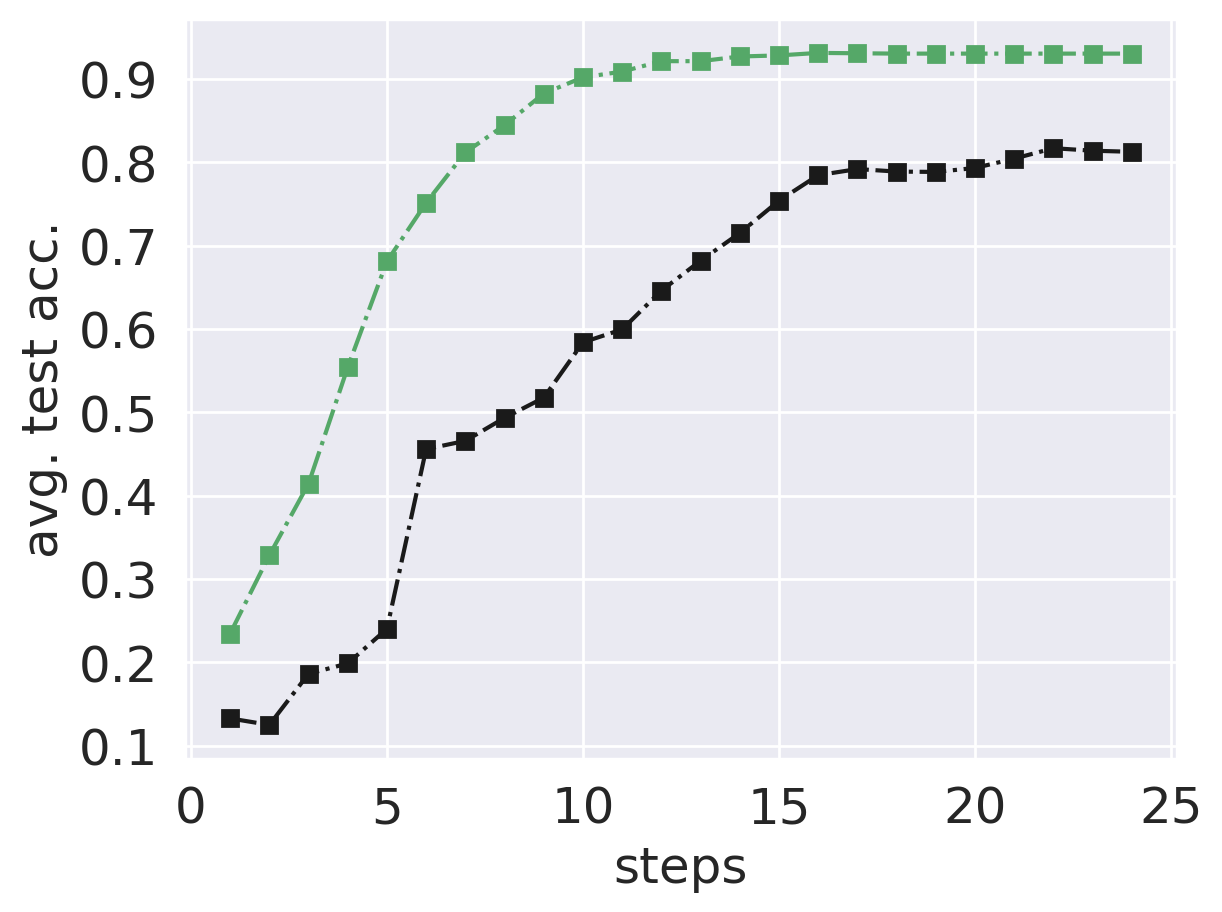}}
    \caption{Time series classification accuracy.}
    \label{fig:ts_cls}
    \end{minipage}
    \hfill
    \vspace{-0.1in}
    \begin{minipage}{0.48\textwidth}
    \centering
    \subfigure[digits]{\includegraphics[width=0.48\textwidth]{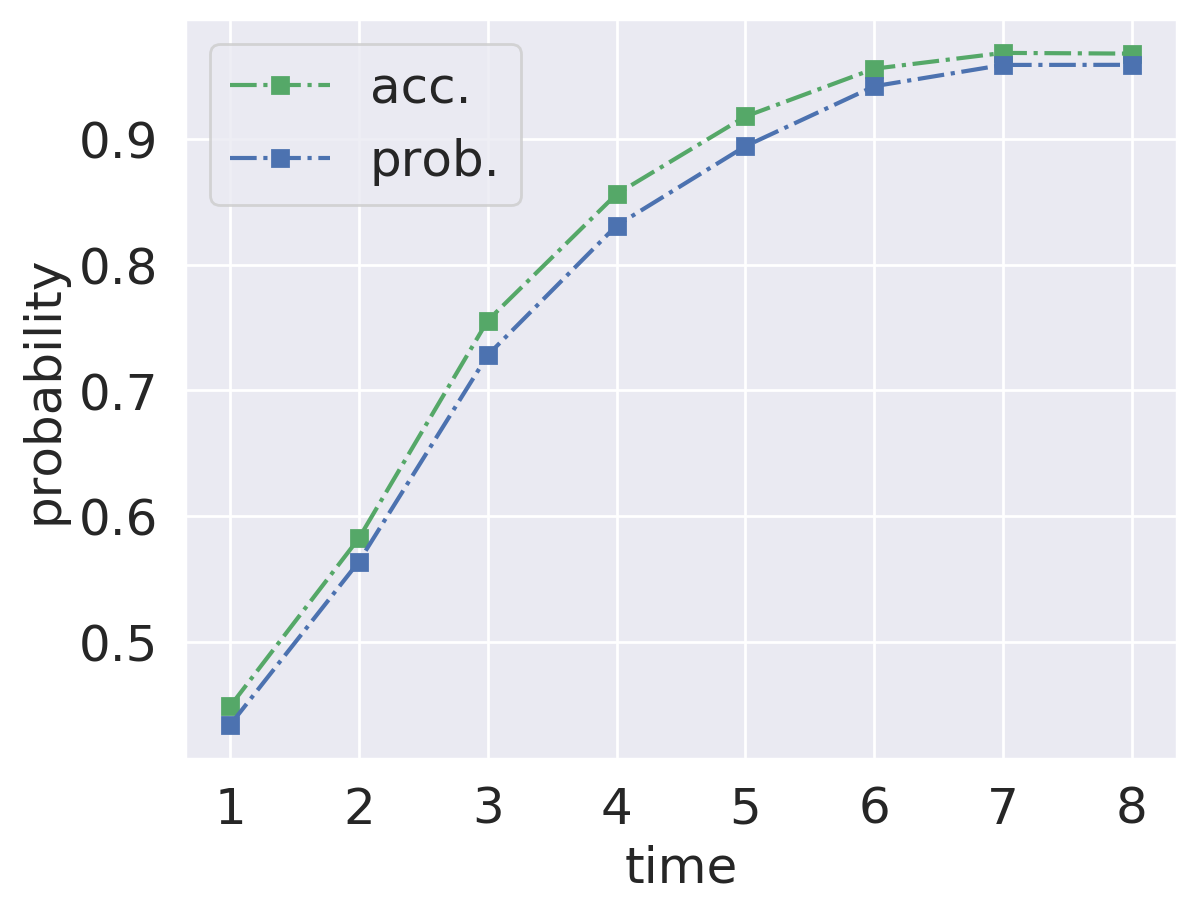}}
    \subfigure[pedestrian]{\includegraphics[width=0.49\textwidth]{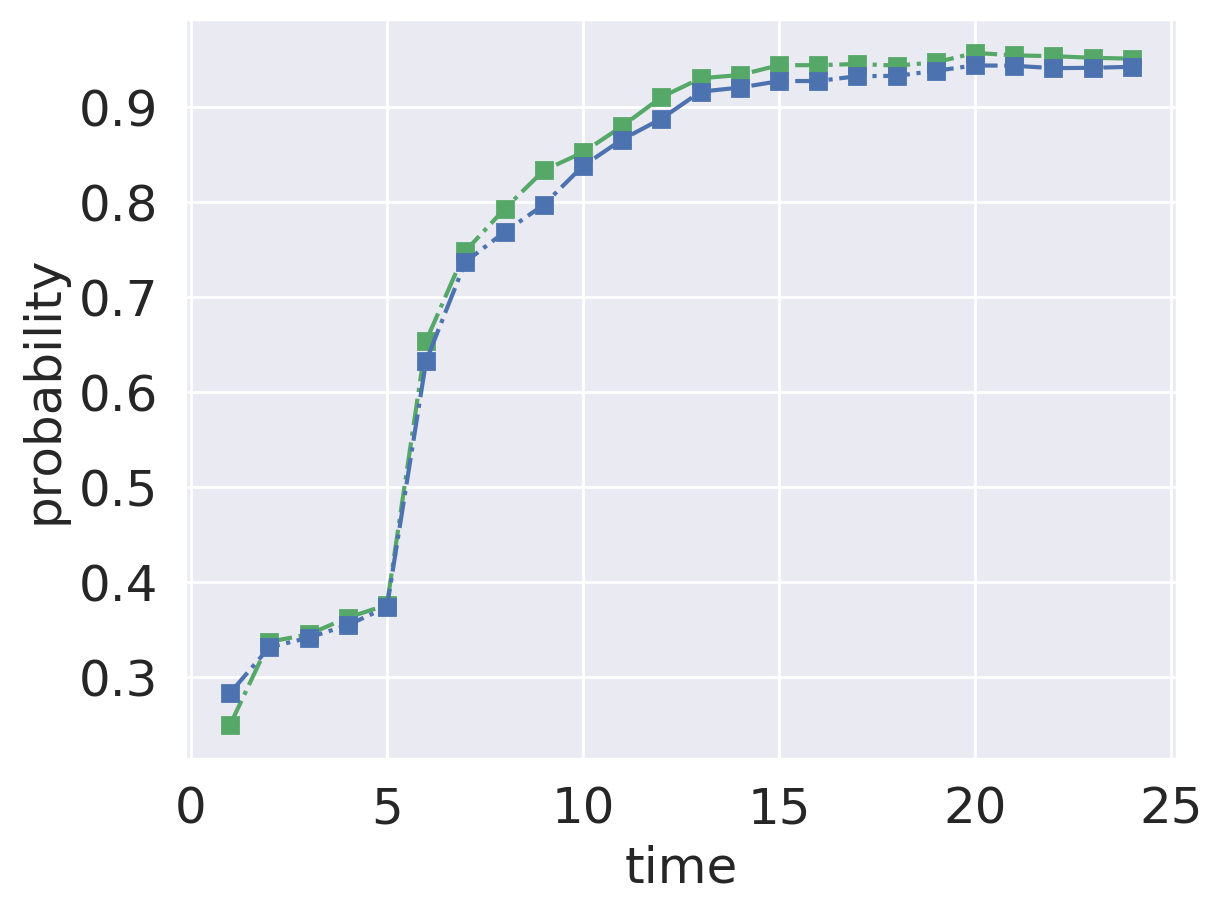}}
    \caption{Accuracy and calibrated probability.}
    \label{fig:ts_uty}
    \end{minipage}
\end{figure}

\vspace{-5pt}
\paragraph{Bayesian Network}
In this section, we utilize the Bayesian network structure to accelerate the acquisition process. We first conduct experiments on two synthetic datasets from Bayesian network repository \cite{bnrepository}, where we know the underlying Bayesian network structure. Data are sampled from conditional Gaussian distributions. Please refer to Appendix~\ref{sec:bn} for data generation process. The results are presented in Fig.~\ref{fig:bn_syn} by comparing our learned BN with the groundtruth BN. We also show the results without using BN for comparison. We notice that utilizing BN does not affect the performance and our learned BN performs equally well compared to the groundtruth cases. Compared to ACFlow DFA without BN, on average our method reduces the candidates set size by 19.6\% and 13.5\% respectively for \textsl{asia} and \textsl{sachs} datasets. We then evaluate the learned BN on UCI datasets for regression in Fig.~\ref{fig:bn_reg}. Compared to the BN learned by the original GS algorithm, our learned BN tends to achieve lower RMSE with the same acquisition budget. Our learned BN also reduces the candidates set size by 28.2\% and 3.6\% for \textsl{housing} and \textsl{whitewine} respectively.

\vspace{-5pt}
\paragraph{Time Series}
In this section, we test our method on time series data. We compare to EDDI by applying the same acquisition policy described in Sec.~\ref{sec:ts}. Figure~\ref{fig:ts_cls} shows the results on two datasets from UEA \& UCR time series classification repository \cite{bagnall16bakeoff}. We can see our model outperforms EDDI by a large margin. Our model reaches the plateau using fewer features, indicating that our model is more efficient for acquisition. We then test the consecutive acquisition case where we use the prediction probability as the stopping criterion. In Fig.~\ref{fig:ts_uty}, we plot the accuracy at each time step (as opposed to acquisition step) and the corresponding prediction probability. We calibrate the probability for each time step separately based on a held-out validation set. We see that after calibration, the estimated probability matches the true accuracy. Thus, one can use the probability as the stopping criterion to make a prediction as soon as possible on a per-instance basis. 
We see that by setting a threshold at 90\%, on average, we would only need to acquire 5.01 and 10.69 time steps for \textsl{digits} and \textsl{pedestrian} datasets respectively, and obtain the requested 90\% accuracy.

\vspace{-5pt}
\section{Conclusion}
\vspace{-5pt}
In this work, we present a framework for dynamic feature acquisition which acquires new features sequentially to improve prediction. We utilize an information theoretic metric, conditional mutual information, to determine which feature to acquire next. The conditional mutual information is estimated by leveraging a flow based generative model, ACFlow, to capture the arbitrary conditional distributions. We conduct experiments in different settings and demonstrate superior performance over strong baselines. To overcome the potential high time complexity for high-dimensional data, we propose to learn the Bayesian network from data using the same ACFlow model. The learned Bayesian network can help reduce the searching space and thus reduce the time complexity. In future work, we will explore this framework in other settings, such as spatial-temporal feature acquisition.

\section*{Broader Impact}
Although human agents routinely reason on instances with incomplete and muddied data (and weigh the cost of obtaining further features), much of machine learning (ML) is devoted to the unrealistic, sterile environment where all features are observed, and further information on an instance is obviated. Thus, the current paradigm in ML is woefully unprepared for the future of automation, which will be highly interactive and able to obtain further information when making predictions. To enable ML for a future of automation that makes decisions in realistic, uncertain environments, we propose to develop dynamic feature acquisition (DFA) methods that can intelligently make assessments on instances in the face of incomplete and costly features. 

The impact of DFA would be broadly felt in a myriad of domains across the sciences and business. For instance, DFA could inform autonomous vehicles when it is possible to accurately make an early prediction based on the limited data collected. In these time-critical applications, DFA has the potential to save lives. DFA could also make a direct impact on applications where the computer is actively interacting with a user to gather information and make an assessment. Examples include surveys and costumer service chat-bots, where DFA would enable such systems to determine the most informative next question and to quickly and accurately make predictions based on the limited answers collected. Related applications of interest are AI education systems that personalize curricula and materials to each student. DFA would enable an education system to quickly assess the student’s current knowledge and abilities with a limited number of questions, which would allow the system to expedite the curriculum and focus on concepts not well understood by the student. DFA would also have a deep impact for applications that collect data out in the field. For instance, in environmental applications agents must physically collect samples from various locations to assess if a region has been contaminated. DFA would not only potentially reduce the number of locations sampled, but may also be used to suggest the most informative locations to query with autonomous agents. DFA would also have impact in medical and health applications. For example, as aforementioned, DFA could sharply reduce the number of invasive biopsies collected on a patient when diagnosing in health-related applications.

Since DFA strives to reduce the number of acquired features and thus the cost of acquisition, a failure of the system would only decrease the efficacy. At worst, it will just degenerate to the case where we acquire all features. 
However, like any other model, DFA might unintentionally learn spurious correlations and biases within the dataset. Thus, we encourage practitioners to carefully design the training set or utilize other debiasing techniques.

\bibliographystyle{unsrt}
\bibliography{main}

\clearpage
\appendix
\section{Proof of Theorem 1}\label{sec:proof}
\begin{proof}
Since Markov Blanket is symmetric, it suffices to prove
$$
I(\mathrm{x}_i ; \mathrm{x}_j \mid \mathrm{x}_{\widebar{\{i,j\}}}) = 0 \iff \mathrm{x}_j \notin MB(\mathrm{x}_i)
$$
From the definition of MB, we know there are three types of nodes could be in MB of a node, i.e., its parents, its children and its spouses. 
\begin{enumerate}
    \item If $\mathrm{x}_j$ is one of the parents of $\mathrm{x}_i$, then $\mathrm{x}_j$ and $\mathrm{x}_i$ are dependent, thus $I(\mathrm{x}_i;\mathrm{x}_j \mid \mathrm{x}_{\widebar{\{i,j\}}}) > 0$.
    \item Similarly, if $\mathrm{x}_j$ is one of the children of $\mathrm{x}_i$, $I(\mathrm{x}_i;\mathrm{x}_j \mid \mathrm{x}_{\widebar{\{i,j\}}}) > 0$.
    \item If $\mathrm{x}_i$ and $\mathrm{x}_j$ are spouses, since their children must be in $\mathrm{x}_{\widebar{\{i,j\}}}$, $\mathrm{x}_j$ and $\mathrm{x}_i$ become dependent conditioned on their children, therefore $I(\mathrm{x}_i;\mathrm{x}_j \mid \mathrm{x}_{\widebar{\{i,j\}}}) > 0$.
\end{enumerate}
Therefore, $\mathrm{x}_j \in MB(\mathrm{x}_i) \Rightarrow I(\mathrm{x}_i ; \mathrm{x}_j \mid \mathrm{x}_{\widebar{\{i,j\}}}) > 0$.

On the contrary, if $I(\mathrm{x}_i;\mathrm{x}_j \mid \mathrm{x}_{\widebar{\{i,j\}}}) > 0$, i.e., $\mathrm{x}_i$ and $\mathrm{x}_j$ are conditionally dependent, which means they are either directly connect (parents or children) or connected by a v-structure (spouses), otherwise there exist a node can make them d-separated.
Therefore, $I(\mathrm{x}_i ; \mathrm{x}_j \mid \mathrm{x}_{\widebar{\{i,j\}}}) > 0 \Rightarrow  \mathrm{x}_j \in MB(\mathrm{x}_i)$.

In all, we have proved that $I(\mathrm{x}_i ; \mathrm{x}_j \mid \mathrm{x}_{\widebar{\{i,j\}}}) > 0 \iff  \mathrm{x}_j \in MB(\mathrm{x}_i)$. 

Since $I(\mathrm{x}_i ; \mathrm{x}_j \mid \mathrm{x}_{\widebar{\{i,j\}}}) \geq 0$, the equivalent contrapositive shows that 
$$
I(\mathrm{x}_i ; \mathrm{x}_j \mid \mathrm{x}_{\widebar{\{i,j\}}}) = 0 \iff \mathrm{x}_j \notin MB(\mathrm{x}_i)
$$
\end{proof}

\section{Arbitrary Conditional Flow}\label{sec:acflow}
ACFlow \cite{li2019flow} belongs to a broader class of model called normalizing flow, where a sequence of invertible (bijective) transformations $q$ are applied on $x$. According to the change of variable theorem, the likelihood for $x$ can be expressed as
\begin{equation}\label{eq:chv}
    p_{\mathcal{X}}(x) = \left| \det \frac{dq}{dx} \right| p_{\mathcal{Z}}(q(x)),
\end{equation}
where $p_{\mathcal{X}}$ and $p_{\mathcal{Z}}$ are likelihood evaluated on input space and latent space respectively.
Typically, latent variables are assumed following a simple base distribution, such as isotropic Gaussian. Since the transformations are invertible, we can generate samples by inverting the transformations over samples from the latent space.

ACFlow extends flow models to learn the arbitrary conditional distributions $p(x_u \mid x_o)$. It extends the change of variable theorem in \eqref{eq:chv} to an arbitrary conditional version:
\begin{equation}
    p_{\mathcal{X}}(x_u \mid x_o) = \left| \det \frac{dq_{x_o}}{dx_u} \right| p_{\mathcal{Z}}(q_{x_o}(x_u) \mid x_o),
\end{equation}
where $q_{x_o}$ refers to the conditional transformations. They propose several conditional transformations and conditional likelihoods based on masking to deal with arbitrary dimensionality of $x_u$ and $x_o$.

Since both $x_u$ and $x_o$ could be an arbitrary subset of $x$, it requires to model an exponential number of conditionals with respect to the dimensionality of $x$. They then train ACFlow in a multi-task fashion, i.e., \emph{all} conditionals are captured by one single model. The multi-task training mechanism could act as a regularizer so that the model can even generalize to unseen combinations of $x_u$ and $x_o$. 

\section{Experiments}
Our models are implemented in Tensorflow 1.0 and trained on a single NVIDIA TITAN Xp GPU. We use Adam optimizer with default hyperparameters throughout the experiments. We train our model for 3000 epochs and utilize early stopping to avoid overfitting. The initial learning rate is set to 0.001 and decays as training.

\subsection{Classification and Regression}\label{sec:cls_reg}
\subsubsection{MNIST dataset}
To reduce the total number of features, we downsample the images to $16 \times 16$. Similar to ACFlow \cite{li2019flow}, we dequantize the pixel values by adding independent uniform noise. The ACFlow model we use here is similar to what they used for MNIST dataset, which contains a stack of conditional coupling transformations and a final conditional Gaussian likelihood layer. The difference is that we also condition on the target variable $y$. We represent $y$ as one-hot vectors and concatenate it with $x_o$ as additional channels. For EDDI \cite{ma2018eddi} baseline, we use their official code and validate the hyperparameters over a held-out validation set. Specifically, we search over the number of layers for both encoder and decoder from 3 to 6 layers, the number of feature maps for each layer from 50 to 512, the dimension of latent code from 10 to 100, the size of set embedding from 10 to 100, etc. We ended up using a architecture with latent code of size 50, set embedding of size 50, four layers of encoder with feature map size 256-128-64-50 and four layers of decoder with feature map size 64-128-256-256. We found that increasing the model capacity does not further improve the performance.

\subsubsection{Synthetic and UCI datasets}
To verify the superiority of DFA over SFA, we build a synthetic dataset where a fixed acquisition order cannot find the optimal subsets. Specifically, we sample from a hierarchical model. We first sample $x_0$ from the uniform distribution and divide the range $[0,1]$ into 9 ranges. If $x_0$ falls into the $i_{th}$ range, we then sample $x_i$ from $\mathcal{N}(w_1*y+w_2*x_0, 0.3)$, otherwise we sample $x_i$ from $\mathcal{N}(0, 1)$. For this dataset, the optimal policy is to first acquire $x_0$ and then acquire the corresponding feature based on its value. Using a fixed acquisition order will need to acquire all features. We generate 20,000 samples and divide them into training, validation and test sets using the ratios 80\%, 10\% and 10\% respectively.
For UCI datasets, we normalize the features to the range $[0,1]$ and split into training, validation and test set using the ratios 80\%, 10\% and 10\% respectively.

The architecture of ACFlow contains four conditional transformation layers (one layer includes one linear transformation, one leaky-relu activation and one coupling transformation) and an autoregressive likelihood module. We similarly validate the hyperparameters for EDDI and select the best one. We found the architecture with a three-layer encoder (200-100-50), a three-layer decoder(50-100-200), 20-dimensional latent code and 20-dimensional set embedding works best for these datasets.

\subsection{Bayesian Network}\label{sec:bn}
The synthetic datasets are generated based on BN structures from Bayesian network repository \cite{bnrepository}. Each node corresponds to a feature. Features are sampled from conditional Gaussian distributions, where the mean of each feature is linear combination of its parents. We set the variance as a constant (0.3). The weights are sampled from Uniform distribution $U(0,1)$. We randomly choose one variable as the target variable $y$.

The ACFlow architecture is the same as we used for regression experiments. We also use the same pretrained ACFlow model to learn the BN structure. The threshold $\epsilon$ is set to 0 throughout our experiments. For comparison, We use Grow Shrink implemented in the BNLearn R package to infer the BN structure.

\subsection{Time Series}
Similar to the UCI datasets, we normalize the features to range $[0,1]$. The \textsl{digits} dataset contains 8 time steps and total 16 features. At each acquisition step, we acquire the corresponding two features at the selected time step.
The \textsl{pedestrian} dataset contains 24 time steps and one feature per time step. 

The ACFlow and EDDI architectures are the same as UCI classification experiments. We set $\alpha$ to 10 for all experiments. For uncertainty calibration, we utilize the validation set and calibrate separately for each time step. We set the number of bins to 10 for the calibration algorithm.

\end{document}